\documentclass[11pt]{article}

\usepackage[final]{acl}
\usepackage{times}
\usepackage{latexsym}
\usepackage[T1]{fontenc}
\usepackage[utf8]{inputenc}
\usepackage{microtype}
\usepackage{inconsolata}
\usepackage{graphicx}
\usepackage{xurl}

\title{GATech at AbjadMed: Bidirectional Encoders vs. Causal Decoders: Insights from 82-Class Arabic Medical Classification}

\author{Ahmed Khaled Khamis \\
  Georgia Institute of Technology \\
  \texttt{akhamis6@gatech.edu}
}

\begin{document}
\maketitle
\begin{abstract}
This paper presents system description for Arabic medical text classification across 82 distinct categories. Our primary architecture utilizes a fine-tuned AraBERTv2 encoder enhanced with a hybrid pooling strategies, combining attention and mean representations, and multi-sample dropout for robust regularization. We systematically benchmark this approach against a suite of multilingual and Arabic-specific encoders, as well as several large-scale causal decoders, including zero-shot re-ranking via Llama 3.3 70B and feature extraction from Qwen 3B hidden states. Our findings demonstrate that specialized bidirectional encoders significantly outperform causal decoders in capturing the precise semantic boundaries required for fine-grained medical text classification. We show that causal decoders, optimized for next-token prediction, produce sequence-biased embeddings that are less effective for categorization compared to the global context captured by bidirectional attention. Despite significant class imbalance and label noise identified within the training data, our results highlight the superior semantic compression of fine-tuned encoders for specialized Arabic NLP tasks. Final performance metrics on the test set, including Accuracy and Macro-F1, are reported and discussed.
\end{abstract}

\section{Introduction}
\textbf{AbjadMed} shared task \cite{gupta2026abjadmed} requires the classification of Arabic medical queries into 82 distinct categories. This high-cardinality classification task has a significant class imbalance, where training samples range from several hundred to as few as seven per category. Furthermore, manual inspection of the provided training data reveals a degree of label noise, where semantically similar categories (e.g., "General Medicine" vs. "Internal Medicine") are sometimes inconsistently assigned. These factors necessitate a system that is not only linguistically nuanced but also highly robust to variance and noise.

A central theme of current Natural Language Processing research is the tension between specialized, fine-tuned bidirectional encoders (such as BERT and its derivatives) \cite{transformers} and large-scale causal decoders (such as the Llama/GPT family) \cite{llama} \cite{gpt}. While the latter have demonstrated remarkable zero-shot reasoning capabilities, their "causal" nature—processing text in a single direction might be suboptimal for capturing the dense semantic boundaries required for 82-class medical categorization. In this paper, we investigate whether the massive parameters and generative pre-training of decoders can outperform task-specific fine-tuned encoders.

To address these challenges, we present our system for medical classification the AbjadMed shared task , \footnote{Code: \url{https://github.com/KickItLikeShika/abjadmed}} the system is designed for high-cardinality classification. Our primary approach leverages a fine-tuned AraBERTv2 encoder \cite{arabert}. We enhance the standard classification architecture with hybrid pooling strategies, concatenating attention \cite{attention-pooling} and mean representations \cite{mean-pooling} to capture both global thematic context and specific medical keywords. To combat class imbalance and label noise, we implement Multi-Sample Dropout \cite{multi-sample-dropout} and Label Smoothing \cite{label-smoothing}, providing a form of internal ensembling that stabilizes the decision boundaries for minority classes.

We contrast our primary system with several exploratory approaches, including a two-stage hybrid re-ranking pipeline using Llama 3.3 70B \cite{llama3} Instruct and feature extraction using Qwen 3 3B \cite{qwen3} embeddings. Our comparative analysis provides evidence that bidirectional encoders remain the superior choice for high-granularity Arabic medical tasks, offering more effective semantic compression than their much larger causal counterparts.

\section{Background}
\subsection{Dataset}
The dataset provided for the shared task consists of Arabic medical queries and their corresponding category labels. The training set contains 27,951 samples distributed across 82 distinct medical categories. The test set comprises 18,634 unlabeled samples.

\begin{figure}
    \centering
    \includegraphics[width=1\linewidth]{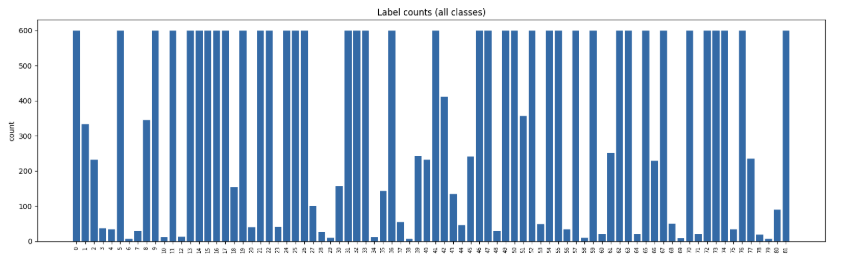}
    \caption{Training set class distribution}
    \label{fig:class_distrib}
\end{figure}

(Figure~\ref{fig:class_distrib}) shows the extreme class imbalance. While majority classes such as Hematological diseases, Addiction, and Neurological diseases are represented by 600 samples each, minority classes such as Biochemistry, Vascular surgery, and In vitro fertilization (IVF) contain as few as 7 samples. This long-tail distribution poses a significant challenge for standard cross-entropy loss, as the model may easily overfit to majority class features while failing to generalize for rarer specialties.

\subsection{Label Noise and Ambiguity}
Beyond the quantitative imbalance, a qualitative analysis of the training data revealed challenges regarding label consistency. During our exploratory phase, we identified several instances of label noise, where the assigned gold-standard label appeared inconsistent with the textual content of the query. Examples include: 1) Queries clearly describing dermatological symptoms were occasionally labeled as General Medicine rather than Dermatological diseases, 2) Mismatch between categories like Sexual Health and Medicinal herbs.
This inherent noise suggests that the dataset contains "soft" boundaries between certain categories. Consequently, we opted for Label Smoothing during training to prevent the model from becoming overly confident in potentially incorrect labels, thereby encouraging better generalization across these ambiguous semantic regions.

\subsection{Data Preprocessing}
\begin{figure}
    \centering
    \includegraphics[width=1\linewidth]{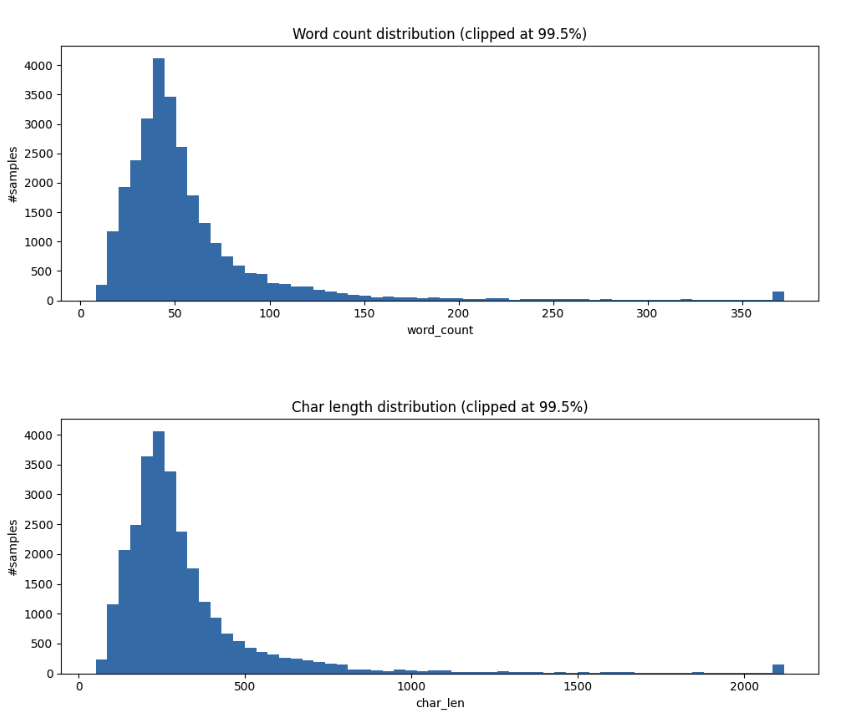}
    \caption{Word and character count distribution clipped at 99.5\%}
    \label{fig:length}
\end{figure}

Given the varied length of medical queries (Figure \ref{fig:length}), we utilized a dynamic padding strategy \cite{padding}. Rather than padding all sequences to a fixed global maximum length, we padded each batch to the length of its longest sequence. This reduced the computational overhead of processing unnecessary padding tokens, allowing for faster iterations and more efficient GPU utilization during the fine-tuning process.

\section{System Architecture}
Our system follows a fine-tuning paradigm where a pre-trained bidirectional encoder is augmented with a specialized classification head and hybrid pooling layers.

\subsection{Encoder Selection}
After benchmarking multiple architectures, we selected AraBERTv2 (bert-base-arabertv02) as our primary encoder. AraBERT is pre-trained on a massive corpus of Arabic text, making it particularly effective at capturing the semantic nuances of medical Modern Standard Arabic (MSA). We utilize the transformer's hidden states as the foundation for our discriminative features.

\subsection{Hybrid Pooling Strategy}
Standard text classification often relies solely on the representation of the [CLS] token \cite{cls-pooling}. However, for medical queries where diagnostic evidence may be scattered across the sequence, a more comprehensive aggregation is required. We implemented a Hybrid Pooling layer that concatenates two distinct representations:
\textbf{Mean Pooling:} Calculates the average of all token embeddings (excluding padding), providing a global thematic summary of the query. 
\textbf{Attention Pooling:} A learnable attention mechanism that assigns importance scores to individual tokens. This allows the model to "focus" on salient medical keywords (e.g., specific symptoms or organ names) while down-weighting irrelevant stop words.
The resulting feature vector $h_{pooled} = [h_{mean} ; h_{attn}]$ doubles the hidden dimensionality ($hidden \times 2$), providing a richer signal to the classification head.

\subsection{Classification Head}
The concatenated features are passed through a dense layer with Layer Normalization \cite{layernorm} and GELU \cite{gelu} activation. To address the dual challenges of class imbalance and label noise, we employ Multi-Sample Dropout.
Unlike standard dropout, which applies a single mask, Multi-Sample Dropout creates five parallel dropout paths with varying rates (0.1 to 0.3) during training. The final classification layer is applied to each of these paths, and the resulting logits are averaged. This technique acts as an internal ensemble, effectively regularizing the model and reducing the variance of predictions, which is critical for the stability of minority class boundaries.

\subsection{Training Methodology}
To preserve the linguistic knowledge of the pre-trained encoder while allowing the classification head to adapt to the 82-class schema, we utilized several advanced optimization techniques:
\textbf{Layer-wise Learning Rate Decay (LLRD):} We applied a decay factor of 0.95, ensuring that layers closer to the input (embeddings) adapt more slowly than the task-specific top layers.
\textbf{Label Smoothing:} We applied a smoothing factor of 0.1 to the cross-entropy loss. This prevents the model from over-fitting to the "hard" labels, which we identified as potentially noisy, and encourages the learning of more flexible decision boundaries.
\textbf{AdamW Optimizer:} Coupled with a Cosine Learning Rate Scheduler, starting at a base learning rate of $2 \times 10^{-5}$ \cite{adamw}.

\section{Experimental Setups}
\subsection{Encoders}
We conducted a systematic evaluation of several Transformer-based encoders, spanning both Arabic-specific and multilingual models.

AraBERTv2 (bert-base-arabertv02) consistently emerged as the top performer. Its specialized pre-training on Arabic-specific corpora allows it to better capture the linguistic nuances of medical Modern Standard Arabic (MSA) compared to multilingual alternatives.
multilingual-E5-large \cite{e5} provided the second-best performance. We attribute its strength to its contrastive pre-training on large-scale text pairs, which yields robust semantic representations across languages.

CamelBERT \cite{camelbert} and mDeBERTa-v3 \cite{deberta} showed competitive performance but were slightly less effective at capturing the specific medical vocabulary of the shared task.
EuroBERT \cite{eurobert} and mmBERT \cite{mmbert} lagged significantly behind, likely due to a lack of representative Arabic medical data in their initial pre-training sets.

\subsection{The Semantic Limitations of Causal Decoders}
We investigated whether the massive parameter count and extensive pre-training of the Qwen 3 3B model could provide superior semantic embeddings compared to smaller BERT-style encoders. We extracted the hidden states from the final layer of the model to serve as static features for a task-specific classification head.

Our results showed that this approach significantly underperformed compared to the fine-tuned AraBERT. We attribute this to the fundamental difference between Causal and Bidirectional Attention:
\textbf{Sequence Bias:} Causal decoders are optimized for next-token prediction, which biases their internal representations towards the sequence history.
\textbf{Semantic Compression:} Unlike bidirectional encoders, which allow every token to attend to the entire context simultaneously, the decoder’s hidden states are "generative" in nature. They fail to compress the full semantic meaning of a medical query into a dense vector suitable for discriminative 82-class categorization. Fine-tuned encoders produce sharper semantic boundaries that are essential for distinguishing between similar medical sub-specialties.

\subsection{Zero-shot Hybrid Re-ranking}
Finally, we explored a two-stage hybrid pipeline. In this setup, AraBERT was used as a candidate generator to propose the Top-15 most probable categories. These candidates, along with the original text, were passed to Llama 3.3 70B Instruct via a structured zero-shot prompt, asking the model to make the final selection. This approach introduced more noise than it resolved. The LLM frequently encountered a "Schema Mismatch": while its choices were often semantically logical (e.g., choosing Dermatology for a skin-related query), they often conflicted with the specific boundaries of the shared task’s 82-class schema (where the correct label might be Skin and Beauty). While the LLM possesses superior general reasoning, the fine-tuned AraBERT encoder more accurately learned the specific intent and labeling conventions of the task's annotators.

\section{Results}
We evaluated the performance of our system on the official competition public test set. Given the extreme class imbalance across the 82 categories, we utilize the Macro-F1 score as our primary performance metric, as it provides an unweighted average that accounts for the model's ability to generalize to minority classes. Table~\ref{tab:results} summarizes the performance of our primary architecture compared to our exploratory benchmarks and the hybrid re-ranking pipeline. Our fine-tuned AraBERTv2 configuration, incorporating hybrid pooling and multi-sample dropout, achieved the highest performance. Interestingly, the addition of a large-scale decoder for re-ranking (Llama 3.3 70B) resulted in a performance degradation, highlighting the "schema mismatch" between general reasoning models and task-specific classification boundaries.

\begin{table}[ht]
\centering
\begin{tabular}{lc}
\hline
\textbf{Model Configuration} & \textbf{Macro-F1} \\ \hline
\textbf{AraBERTv2} & \textbf{0.3934}   \\
multilingual-E5-large        & 0.3804            \\
AraBERTv2+Llama 3.3 70B & 0.3035   \\
CamelBert                  & 0.3603            \\
Qwen 3 3B (Feature Extraction) & 0.1278         \\ \hline
\end{tabular}
\caption{System performance on the official test set. The primary AraBERTv2 configuration utilizes our proposed hybrid pooling and multi-sample dropout head.}
\label{tab:results}
\end{table}

\section{Discussion and Limitations}
\subsection{The Encoder Advantage in High-Cardinality Tasks}
Our results reinforce the hypothesis that for high-cardinality classification within a specific domain, discriminative fine-tuning on a bidirectional encoder is more critical than the general reasoning capabilities of large causal decoders. While models such as Llama~3.3~70B contain extensive medical knowledge, they are not aligned with the fine-grained annotation boundaries of this shared task. In contrast, fine-tuning allows AraBERT to sharpen semantic decision boundaries and distinguish between medically similar but distinctly labeled categories (e.g., Alternative medicine vs. Medicinal herbs).

\subsection{The Inefficacy of Generative Embeddings}
A key finding of this study is the poor performance of feature extraction from causal decoders such as Qwen~3~3B and Llama~3~8B. We observe that causal attention can be suboptimal for sequence-level semantic compression, as the final token embedding often reflects recent sequence history rather than the full input. Bidirectional encoders, through all-to-all attention, allow medical keywords appearing anywhere in the query to contribute equally to the final representation, which is especially important when diagnostic cues occur early in long, conversational medical queries.

\subsection{Future Work}
While our modeling experiments combine several training and architectural techniques, a more fine-grained analysis remains valuable. Future work should include ablation studies that evaluate the individual impact of hybrid pooling, multi-sample dropout, and label smoothing by introducing each component independently. Such controlled ablations would provide clearer insight into the contribution of each technique.

\section{Conclusion}
We presented a system for medical text classification in the AbjadNLP shared task, addressing the challenges of extreme class imbalance across an 82-class label space. Our approach, based on fine-tuning an AraBERTv2 encoder with hybrid pooling and multi-sample dropout, achieved strong performance compared to multilingual encoders and large causal decoders.

\bibliography{custom}
\end{document}